\documentclass[journal]{IEEEtran}
\ifCLASSINFOpdf
   \usepackage[pdftex]{graphicx}
	 
	\usepackage{amsmath}
\usepackage{amssymb}    
\usepackage{caption}

\else
  \usepackage[dvips]{graphicx}
\fi

\usepackage{tabularx}
\usepackage{amsmath}
\usepackage{amssymb}
\ifCLASSOPTIONcompsoc
  \usepackage[caption=false,font=normalsize,labelfont=sf,textfont=sf]{subfig}
\else
  \usepackage[caption=false,font=footnotesize]{subfig}
\fi
\begin{document}
%
\title{On the Use of Client-Specific Information for Face Presentation Attack Detection Based on Anomaly Detection}
%
%
%
\author{Shervin~Rahimzadeh Arashloo~\IEEEmembership{}
				and~Josef~Kittler,~\IEEEmembership{}

\thanks{S.R. Arashloo is with the CVSSP, Department of Electronic Engineering, University of Surrey, Guildford, Surrey, GU27XH UK e-mail: S.Rahimzadeh@surrey.ac.uk.}
\thanks{J. Kittler is with the CVSSP, Department of Electronic Engineering, University of Surrey, Guildford, Surrey, GU27XH UK e-mail: J.Kittler@surrey.ac.uk.}
\thanks{}}

%
%

\markboth{2018}%
{Arashloo \MakeLowercase{\textit{et al.}}: Title}
%



\maketitle

\begin{abstract}
The one-class anomaly detection approach has previously been found to be effective in face presentation attack detection, especially in an \textit{unseen} attack scenario, where the system is exposed to novel types of attacks. This work follows the same anomaly-based formulation of the problem and analyses the merits of deploying \textit{client-specific} information for face spoofing detection. We propose training one-class client-specific classifiers (both generative and discriminative) using representations obtained from pre-trained deep convolutional neural networks. Next, based on subject-specific score distributions, a distinct threshold is set for each client, which is then used for decision making regarding a test query. Through extensive experiments using different one-class systems, it is shown that the use of client-specific information in a one-class anomaly detection formulation (both in model construction as well as decision threshold tuning) improves the performance significantly. In addition, it is demonstrated that the same set of deep convolutional features used for the recognition purposes is effective  for face presentation attack detection in the class-specific one-class anomaly detection paradigm.
\end{abstract}


%
\IEEEpeerreviewmaketitle

\section{Introduction}
\IEEEPARstart{A}{lthough} biometrics systems have witnessed an increase in their popularity in the past decades, their reliability is seriously challenged by spoofing attacks where an unauthorised subject tries to access the system by presenting fake biometric data. In the case of face recognition systems, spoofing attacks generally appear as print attacks or replay attacks.

During the past couple of years, a variety of different face presentation attack detection (PAD) approaches have been proposed, achieving impressive performance on benchmarking data sets. The progress made mainly owes to two factors: i) the design and deployment of more effective representations which can better capture the differences between real and fake biometrics traits, and ii) using more powerful two-class classifiers. The majority of the approaches to face spoofing detection proposed in the literature formulate the problem as a two-class classification problem and then try to learn a suitable classifier to discriminate between the real accesses and spoofing attempts. Despite the huge advances made in this direction, currently the face PAD methods do not perform robustly and lack the ability to generalise to more realistic application scenarios, where the type of presentation attack cannot be anticipated, and may take a completely new form. The drawbacks of the common two-class formulation include \cite{7984788}:
\begin{itemize}
\item Difficulties in learning an effective decision boundary for classification due to the multi-modal nature of spoofing attack data;
\item Difficulties in augmenting the training set, as attack data is quite complicated and laborious  to collect and, at the same time, it cannot cover all possible unforeseen attacks. While it is possible to enlarge the size of real access data, increasing solely this category of training data  would result in a progressively deteriorating training set imbalance.
\item Lack of generalisation capabilities of the learned two-class systems to accommodate novel attack types.
\end{itemize}
Different approaches have been proposed in the literature to counteract one or more of these shortcomings with various degrees of success. Among others, in order to partly compensate for some of the aforementioned inadequacies, the work in \cite{7984788} formulated the face PAD problem as one of anomaly detection, where the real access data was considered as \textit{normal} and the spoofing attacks were presumed to be \textit{anomalous} observations deviating from normality. Such a one-class formulation offers a number of desirable properties such as:
\begin{itemize}
\item Insulation from the undesirable effects of the spoofing data diversity on the performance, as only normal data is used to build the model;
\item Since only real-access data is required for training, the training set can more easily be extended;
\item Anomaly-based methods are better equipped to detect previously unseen, completely novel attack samples.
\end{itemize}
In order to measure the capacity of one-class systems to detect novel forms of attack, previously unseen by the system, extensive evaluations on different datasets in an \textit{innovative attack evaluation} scenario (where the systems were exposed to novel unseen attack types) were conducted in \cite{7984788}. It was observed that the one-class approaches may perform as well as their two-class counterparts. In a later study, the work in \cite{Nikisins_ICB2018_2018/LIDIAP} followed a similar one-class anomaly detection approach for face PAD using a Gaussian mixture model-based anomaly detector, which exhibited good generalisation properties on novel types of attacks on an aggregated database. Motivated by these observations and the desirable properties of the one-class formulation, the current study follows the same one-class anomaly-based approach with distinctive contributions outlined in the next subsection.

\subsection{Contributions}
An aspect of classifier design which has been unexplored in \cite{7984788,Nikisins_ICB2018_2018/LIDIAP} is the use of client-specific information. Although in any spoofing detection system the representations used are selected in a way that they capture the intrinsic differences between real accesses and spoofing attempts, such an approach does not rule out the possibility that the features used can be affected by the specific characteristics of each individual client \cite{7031941}. From this perspective, the majority of the work on face spoofing detection, including \cite{7984788,Nikisins_ICB2018_2018/LIDIAP}, can be considered as \textit{client-independent} approaches. A client-independent face PAD approach assumes that the relevant information comes from either a real access or the attack class, whereas a client-specific method assumes that the constructed representations are additionally influenced by the identities of the subjects. In \cite{7031941}, it is shown that the client identity information can be deployed to devise two-class classifiers which can make better discrimination between the real accesses and spoofing attacks. The use of identity information in a PAD system is justifiable from the point of view that anti-spoofing mechanisms are designed to guard biometrics systems against spoofing attacks and hence work in conjunction with them. As a result, any identity information available to a recognition system is readily accessible by the spoofing detection module. Other work in \cite{7041231,7820995} have also made use of client identity information to design client-specific two-class countermeasures to face spoofing attacks.

Motivated by the aforementioned observations, it is first shown in this work that the anomaly-based approaches to face spoofing detection can benefit from client identity information to improve performance. Although the use of client specific information in face recognition has been studied extensively \cite{6905848,Arashloo2017,7494969,THANHTRAN2017131,7965980,iet:/content/journals/10.1049/iet-bmt.2017.0081}, the deployment of such information for face spoofing detection in the two-class formulation framework has been limited to just a few studies \cite{7031941,7041231,7820995}. This work advocates the use of such information in a one-class anomaly-detection paradigm and builds effective one-class face PAD mechanisms. The identity information for face PAD is deployed in two stages: (i) instead of a single classifier applicable to all subjects, a separate client-specific one-class classifier is designed for each individual enrolled in the dataset; (ii) using score distributions of each user, subject-specific decision thresholds are determined to make the final decision. It is shown that the use of client-specific information, both in the model construction and in setting a decision threshold, improves the detection performance of one-class spoofing detection systems by a large margin.

Second, inspired by the recent success of deep neural networks and in particular deep convolutional neural networks (CNNs), the proposed one-class approaches are fed with representations obtained from deep pre-trained CNN models. The models employed are either pre-trained for the general object classification purposes or specifically tuned for face recognition. In this respect, a further contribution of the current study is a comparative evaluation of the applicability of different deep CNN models designed for recognition purposes to the problem of   face spoofing detection based on the one-class face PAD formulation. This aspect of the study is particularly important, as the use of similar features for both recognition and spoofing detection engines leads to a very interesting conclusion of practical significance, namely, that the features used  for spoofing detection and recognition can be shared. This can simplify the design of face biometric systems.

A comparison between different well-known deep CNNs
A comparison between different one-class classifiers
A classifier fusion approach

The rest of the paper is organised as follows: In Section \ref{lit}, we review the relevant literature. Section \ref{meth}, introduces the proposed class-specific anomaly detection framework for face spoofing detection. This discussion is then followed by a description of different generative as well as discriminative one-class classifiers and of the deep CNN models examined. In Section \ref{exp}, the results of an experimental evaluation of the proposed class-specific anomaly detection approach are provided and compared with client-independent methods. Finally, conclusions are drawn in Section \ref{conc}.

%
%
%
%


\section{Related Work}
\label{lit}
Face spoofing countermeasures can broadly be  classified into hardware-based and software-based methods \cite{edmunds:tel-01576830}. While software-based methods process the data collected from a typical authentication sensor, hardware-based methods use additional hardware for anti-spoofing. The hardware-based solutions typically rely on liveness measurements (e.g. employing a specific sensor to detect attributes of living bodies), attack specific detection methods (such as depth measurements against photo and video attacks) or challenge-response mechanisms (requiring the user to respond to random requests). Software-based methods on the other hand use different attributes of an image sequence along with different classifiers to detect presentation attacks. Different descriptors used for this purpose include texture, motion, frequency, colour, shape or reflectance, while the two-class classifiers employed include discriminant, regression, distance metric or other heuristic methods.

Among the cues conveyed by an image/image sequence, texture is probably the one most frequently  used for spoofing attack  detection \cite{6612968,6313548,6117592,6847399,7163625,6612955}. The use of texture is based on the assumption that face presentation attacks produce certain texture patterns which do not exist in real access attempt data. Motion-based methods constitute another group where typically two different ways of exploiting motion are considered. The first approach focuses on intra-face variations, such as facial expressions, eye blinking, and head rotation \cite{4409068,4563115,KOLLREIDER2009233,FENG2016451}, while the other alternative is to assess the consistency of the user with the environment \cite{6612968,6485156,7185398}. A different category of methods is constituted by  frequency-based countermeasures proposed to detect certain image artefacts in 1D or 2D Fourier transform from either a single image \cite{6199760} or an image sequence \cite{6382760,7017526,7185398}. Despite the fact that colour does not remain constant due to inconsistencies in imaging conditions, certain colour attributes have also been used to discern attacks from real accesses in a different group of methods \cite{6117592,6976921,7351280}. Another category uses shape as a source of information to deal with some presentation attacks \cite{6612957}. There also exist methods \cite{6622704,KOSE2014779,10.1007/978-3-642-15567-3_37} which use reflectance information for attack detection, based on the assumption that real and attack attempts behave differently under the same illumination conditions.

Along with the use of different cues for attack detection, a variety of two-class classifiers have also been examined for face PAD. Discriminant classifiers constitute one such group of methods where Support Vector Machines are the most commonly employed technique \cite{6117510,6612968,6712690,6180283,10.1007/978-3-642-37410-4_11,KOSE2014779,6612957,7031384}. Other works have also examined the linear discriminant analysis for attack detection \cite{6595861,6976921,6810829}. Other types of classifiers using discriminant procedures include neural networks \cite{FENG2016451}, convolutional neural networks \cite{7029061}, Bayesian networks \cite{6117509} as well as  Adaboost \cite{6613026}. Another group includes the regression-based methods which try to map the input descriptors directly onto their class labels \cite{6612968,6485156,6116484,6382760,7041231}. There also exist methods based on learning a distance metric with the goal of measuring dissimilarities among samples \cite{10.1007/978-3-642-21605-3_19,6117509,6317336}. In addition to the methods mentioned above, different heuristics have also been used for classification in face PAD systems \cite{4409068,4563115,KOLLREIDER2009233,7056504}.

The majority of the work on face spoofing detection assumes that the relevant information for the detection of an attack is independent of the class identity of the data. Accordingly,  the systems are typically designed in a client-independent fashion. However, it has been observed that the representations used for the detection of spoofing attempts are invariably affected by client-specific attributes \cite{7031941}. Drawing on this observation, the work in \cite{7031941} studies how much  client-specific information is contained within features and its effect on the performance of different systems. Using such information, two client-specific anti-spoofing solutions, one generative and the other discriminative are built. The methods proposed outperformed the client-independent methods by a large margin while demonstrating better generalisation capabilities to unseen types of attacks. Other work \cite{7041231} proposed a person-specific anti-spoofing approach using a classifier specifically trained for each subject in an attempt to dismiss the interferences among the subjects. A subject domain adaptation method was then applied to synthesise virtual features, making it possible to train individual face anti-spoofing classifiers. In a different study \cite{7820995}, the face PAD problem is addressed by modelling radiometric distortions involved in the recapturing process. Having access to the enrolment data of each client, the exposure transformation between a test sample and its enrolment counterpart is estimated. A compact parametric representation is then proposed to model the radiometric transform and employed as features for classification.

Although the use of client-specific information in two-class models has led to some improvements, as noted earlier, a common drawback of these two-class approaches is their insufficient capacity to generalise to spoofing attempts of different nature. The detection of novel attack types is particularly challenging, making it impossible to predict the performance of an anti-spoofing technique in real-world scenarios. On the other hand, as it is impossible to foresee all possible attack types and cover them in the database, one-class approaches, modelling only the real-access data present a promising direction towards detection of unseen attack types \cite{6909967}.

\section{Anomaly Detection}
\label{meth}
\subsection{Background}
Anomalies are typically known as set of patterns/conditions which are different in some way from the majority of observations considered as normal. In this respect, anomaly detection is a problem of identifying items, events or observations which do not conform to an expected behaviour or condition. Anomaly detection finds use in a wide variety of applications such as fault detection in safety critical systems, intrusion detection, fraud detection, insurance, health care, surveillance, etc. Anomalies are referred to in the literature with different terminology, including outliers, exceptions, peculiarities, surprises, outliers, novelties, noise, deviations, exceptions, discordant observations, peculiarities, aberrations, contaminants, etc. Among these, anomalies and outliers are the two terms which are the most commonly used interchangeably.

The common notion of anomaly as an outlier from some known class representing normality is referred to as point anomaly in the literature. A general categorisation of anomaly detection methods applicable to point anomaly detection methods considers them to be of either generative or non-generative type \cite{6636290}. While for generative methods there exist a model for generating all observations, non-generative approaches lack a transparent link to the data. The non-generative methods are best represented by discriminative approaches which try to identify the class identity of an observation by partitioning the feature space. The construction of an anomaly detection mechanism can be based on normal data or on both normal and anomalous observations. The merits of using only normal data has been studied in \cite{1262509}. In this work, both generative and discriminative approaches are examined for face spoofing detection in the context of the one-class anomaly detection framework, where only normal (real-access) data is used for training.

\subsection{Class-Specific Anomaly Detection}
Previous studies formulating face spoofing detection as an anomaly detection problem \cite{7984788,Nikisins_ICB2018_2018/LIDIAP} considered real-access data as the \textit{normal} observations and  spoofing attacks as \textit{anomalies}. The examination of different one-class classifiers revealed the merits of such an approach, particularly in the case of an \textit{unseen} attack evaluation scheme. In spite of such an appealing characteristic, the one-class approaches in \cite{7984788,Nikisins_ICB2018_2018/LIDIAP} implicitly make the assumption that the useful information conveyed by the chosen representations for categorisation of a pattern was independent of the client identity. In the case of the two-class formulation of the face PAD problem, this assumption has been re-evaluated in different studies \cite{7031941,7041231,7820995}, with the conclusion that, using client-specific information by virtue of training different client-specific face spoofing detection classifiers led to significant improvements in the system performance.

This work advocates a similar client-specific spoofing detection mechanism but in an anomaly-based paradigm. In order to train a class-specific anomaly detection model, the class labels are required both during the training as well as the operation phase of the system. As discussed in \cite{7031941}, such information is readily available to a face spoofing detection engine as it would work in conjunction with a face recognition system. More specifically, the enrolment data of each subject in the recognition system can be employed to built a subject-specific spoofing detection model. In the operation phase of the spoofing detection system, the class identity information is accessible from the face verification or identification engine. While in the face verification case a test subject claims an identity, in an identification scenario the test image is compared against several models stored in the gallery, whose identities are known. In both cases, the identity of the target class is known and can be utilised by the spoofing detection system. In summary, in the current work, during the training phase of the face spoofing detection system, a separate one-class anomaly detection classifier is trained for each subject using the enrolment data of the corresponding client while in the operation phase, the test sample is matched against the model of the claimed subject. The construction of a subject-specific classifier in this work benefits from the client identity information at two levels. First as noted earlier, only the enrolment data of the subject under consideration is used to build a one-class classifier. Second, by analysing the score distributions of each subject, a subject-specific threshold is determined for the final decision making process. The one-class classifiers examined in this work, including both, discriminative as well as generative approaches, are explained next.

\subsection{One-Class SVM (OCSVM)}
The one-class SVM proposed in \cite{doi:10.1162/089976601750264965} is known to be an effective discriminative approach for the problem of one-class classification aiming at detecting samples which are dissimilar to the majority of the dataset. The one-class SVM method in \cite{doi:10.1162/089976601750264965} attempts to learn a decision boundary that achieves the maximum separation between the points and the origin following a quadratic program formulation. For this purpose, the method in \cite{doi:10.1162/089976601750264965} uses an implicit transformation function defined by a kernel to map input data into a high dimensional feature space. It finds a hyperplane which is maximally distant from the origin, separating the data from the origin. To this end, a binary function is proposed which returns +1 in regions containing the data and -1 elsewhere. For a test sample, the value of the decision function is evaluated to determine which side of the hyperplane it falls on in the feature space.

\subsection{One-Class SRC (OCSRC)}
The sparse representation based classification (SRC) method \cite{4483511,7102696} is known to be a valuable generative approach for classification. The SRC method assumes that a test sample approximately lies in the linear span of training samples and then represents it as a sparse linear combination of the training samples. Different algorithms are available to solve the sparse representation problem. For the L1-minimisation problem (imposing an L1 cost on the sparse coefficients), relatively efficient methods exist \cite{4655448,doi:10.1093/imanum/20.3.389}. From a classification perspective, one can classify the test sample based on the reconstruction error using the sparse coefficients corresponding to a target class. In a one-class problem, the reconstruction residual of the normal class is used as a dissimilarity criterion.

\subsection{Mahalanobis Distance (MD)}
As a baseline method, in this work it is assumed that the representations obtained from a real-access sequence follow a single-mode Gaussian distribution. Despite being a simplistic assumption, it is found to serve as a good baseline approach. Once the parameters characterising the normal distribution are estimated using the real-access (normal) samples, testing for normality entails computing the Mahalanobis distance of a test pattern to the mean of the normal class.

\subsection{Deep Representations}
Motivated by the recent success of deep networks and in particular deep convolutional networks (CNNs), in this work deep pre-trained CNN models are used to derive representations from an image sequence. The chosen CNN models include a deep pre-trained model for the general object recognition problem as well as a model specifically tuned for face recognition purposes. This choice is made to assess the applicability of deep pre-trained CNN models to the face spoofing detection problem in the proposed client-specific one-class framework. In particular, as illustrated in the experimental evaluation section, deep pre-trained models which are specifically tuned for face recognition can be directly employed to detect spoofing attacks. This has important implications as it simplifies face biometrics system design by unifying the feature extraction stages for both recognition and PAD units. A description of the deep CNN models utilised in this work is provided next.

\subsubsection{GoogLeNet}
GoogLeNet \cite{7298594} is a deep convolutional neural network based on the inception model. GoogLeNet achieved the state-of-the-art result for classification and detection in the ImageNet Large-Scale Visual Recognition Challenge 2014 (ILSVRC14) \cite{ILSVRC15}. In this model, following a carefully crafted design, the depth and width of the network was increased compared to the previous networks while keeping the computational budget constant. GoogLeNet is a 22-layer deep network, the quality of which is validated in the context of classification and detection problems.

\subsubsection{VGGFace}
VGGFace \cite{Parkhi15} is a deep CNN model based on the VGG model, comprised of 11 blocks, each containing a linear operator followed by one or more non-linearities such as ReLU and max pooling. The first eight blocks are convolutional, while the last three blocks are fully connected. In this network, the convolution layers are followed by a rectification layer. The model is trained on a very large scale dataset of 2.6M images from over 2.6K subjects. It achieved competitive results on the LFW \cite{LFWTech} and YTF \cite{5995566} face benchmarks.

\section{Experimental Evaluation}
\label{exp}
The aim of the experiments described in this section is to evaluate the performance of different client-specific one-class face PAD methods and and compare it to the client-independent one-class approaches in an unseen attack scenario. As noted earlier, an essential pre-requisite to build a client-specific face anti-spoofing system is the availability of enrolment data for each client in the database. However, the majority of face-spoofing data sets currently in use, including the CASIA FASD \cite{6199754}, MFSD \cite{7031384} and NUAA \cite{10.1007/978-3-642-15567-3_37} lack enrolment data. Although it is possible to violate the protocols of these data sets and use parts of real access data as the enrolment data, in such a case, the dedicated enrolment data and the test samples would come either from a single session or even worse from a single video which would make the data highly correlated and the results strongly biased. In this study we use the Replay-Attack dataset \cite{6313548}, which is a widely used benchmark that provides enrolment data for each client, in addition to the training, development and test sets.

\subsection{The Replay-Attack Data Set}
The Replay-Attack database \cite{6313548} includes video recordings of real-access and attack attempts of fifty different individuals. For each subject, a number of videos in two different conditions, controlled and adverse, were recorded using an Apple MacBook laptop.  The same illumination conditions and background settings were used for real access and presentation attack video recordings.  A 12.1 Megapixel Canon camera and a 3.1 Megapixel iPhone 3GS camera were used to capture two high-resolution images of each subject. Attacks were realised in one of the following scenarios: (1) Print: where hard copies of the high-resolution digital photographs printed on plain A4 paper are displayed; (2) Mobile: where photos and videos taken with the iPhone are displayed on the iPhone screen; and (3) Highdef: where the high resolution digital photos and videos are displayed using an iPad screen. Each attack video is captured in two different modes: (1) Hand-held: where the attack media or device is held by an operator; and (2) Fixed-support: where the attack device is fixed on a support.

The video recordings in this database are divided randomly into three subject-disjoint subsets for training, development and testing purposes. While the test set is solely used to report error rates and to generate the performance curves, the training set is designated for model training and the development set is typically used to set system parameters. In addition to the training, development and test sets, the Replay-Attack database provides an extra set of 100 videos recorded in a separate session that correspond to the enrolment data for each of the fifty subjects.

The performance of a spoofing detection system is commonly reported in terms of Half Total Error Rate (HTER), which is half of the sum of the False Rejection Rate (FRR), and the False Acceptance Rate (FAR) on the test set. When reporting HTER, the decision threshold is set such that FRR and FAR on the development set are equal. In addition, the performance can also be presented using the Receiver Operating Characteristic (ROC) curve, which plots the True Positive Rate (TPR) versus the False Positive Rate (FPR) for different values of the decision threshold. The Area Under the ROC Curve (AUC) can be used as an indicator of the average performance of a system irrespective of a specific decision threshold.

\subsection{Systems Evaluated}
Using a combination of the deep CNN's and the one-class classifiers introduced earlier, different generative methods evaluated in this work are as follows:
\begin{itemize}
\item $OCSRC+GoogLeNet_{spe}$:\\
The client-specific one-class SRC classifier operating on GoogLeNet representations\\
\item $OCSRC+GoogLeNet_{ind}$:\\
The client-independent one-class SRC classifier operating on GoogLeNet representations\\
\item $OCSRC+VGGFace_{spe}$:\\
The client-specific one-class SRC classifier operating on VGGFace representations\\
\item $OCSRC+VGGFace_{ind}$:\\
The client-independent one-class SRC classifier operating on VGGFace representations\\
\item $MD+GoogLeNet_{spe}$:\\
The client-specific Mahalanobis distance classifier operating on GoogLeNet representations\\
\item $MD+GoogLeNet_{ind}$:\\
The client-independent Mahalanobis distance classifier operating on GoogLeNet representations\\
\item $MD+VGGFace_{spe}$:\\
The client-specific Mahalanobis distance classifier operating on GoogLeNet representations\\
\item $MD+VGGFace_{ind}$:\\
The client-independent Mahalanobis distance classifier operating on GoogLeNet representations\\
\end{itemize}
The discriminative approaches evaluated are as follows:
\begin{itemize}
\item $OCSVM+GoogLeNet_{spe}$:\\
The client-specific one-class SVM classifier operating on GoogLeNet representations\\
\item $OCSVM+GoogLeNet_{ind}$:\\
The client-independent one-class SVM classifier operating on GoogLeNet representations\\
\item $OCSVM+VGGFace_{spe}$:\\
The client-specific one-class SVM classifier operating on VGGFace representations\\
\item $OCSVM+VGGFace_{ind}$:\\
The client-independent one-class SVM classifier operating on VGGFace representations\\
\end{itemize}

\subsection{Implementation Details}
\begin{figure*}[t]
\centering
\includegraphics[width=4.5in]{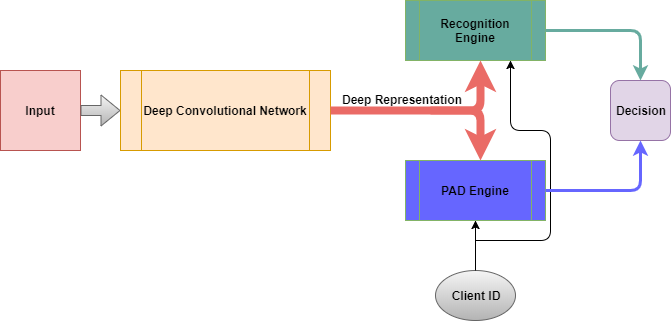}
\caption{The proposed approach for client-specific anomaly-detection based PAD.}
\label{system}
\end{figure*}
In the following experiments, in order to minimise the effect of background on the detection performance, only the face regions are used in each sequence. For this purpose, the coordinates of the faces provided along with the Replay-Attack dataset are used to crop out a face from the entire image in each frame. In the case of a missing bounding box for a frame, the coordinates of the last detected face in the same sequence are used instead. Next, the CNN models described earlier are used as mechanisms for feature extraction. To this end, once the final fully-connected layers of each model performing the classification are removed,  the models are applied to the facial regions of each individual frame. This results in feature vectors of 4096 elements for the VGGFace network and 1024-element feature vectors for GoogLenet, per each frame.

For the construction of OCSVM models, once the feature vectors are L2-normalised, MatlabR2017b is used to build the one-class SVM classifiers using a radial basis function kernel. The normalization parameter in the kernel is determined automatically. The Mahalanobis distance is implemented as a Euclidean distance in the PCA space, the dimensionality of which is determined by retaining $99\%$ of the variance in the data. Each feature vector in this case is mapped to a lower dimensional space using the leading eigenvectors and then divided by the square root of the corresponding eigenvalues. 

For building the OCSRC models, the sparse coefficient are obtained via the Homotopy-based algorithm of \cite{4655448}, where deep representations obtained from the enrolment/training frames are used as dictionary elements. While for the OCSVM and MD classifiers we use the feature vectors corresponding to all the enrolment/training frames, in the case of OCSRC, due to the high computational complexity of the method, the number of dictionary elements is set to $10\%$ of the total available samples. An analysis of the effect of the training sample size on system performance is conducted in the experimental evaluation section. 

The \textit{enrolment} data available for each client in the data set is used for training and to build the class-specific models. In contrast, the real-access \textit{training} data is used for the construction of class-independent models. In the evaluation phase, a query is matched only against the claimed client model. 
It should be reiterated that in both, class-specific and class-independent cases, only the real access data is used to build a one-class anomaly detection model. As a result, the evaluation scheme is \textit{unseen} in the sense that none of the attack types is seen during the training phase of the system.

The main purpose of the experiments described  in this section is to demonstrate the merits of using client-specific information in the selected CNN feature space in conjunction with the anomaly detection paradigm.  As such, the construction of the one-class face spoofing detection mechanisms is not aimed to compete with the best performing two-class methods utilising both real access and spoofing data. Instead, the focus here is on improving the performance of one-class anomaly based methods which possess the appealing characteristics listed earlier. Nevertheless, the proposed methods can be compared to the existing approaches in the literature provided that the same evaluation scenario (i.e. unseen attack type) is followed. As a by-product of the experiments, it will be shown that the deep CNN models examined possess the potential to be used directly for feature extraction for face presentation attack detection in a one-class client-specific anomaly detection framework. The proposed approach is depicted in block diagram form in Fig. \ref{system}.

\subsection{Per-Frame Results}
As the employed deep CNN models are applied to each individual frame, it is possible to report the performance of different systems on a per-frame basis. The criteria for the evaluation of various methods are the  AUC, EER and HTER measures. AUC indicates the area under the TPR vs. FPR curve on the test set while EER denotes the equal error rate (FNR=FPR) on the development set. HETR is the mean of FNR and FPR on the test set when the threshold is set to the EER point on the development set. Table \ref{EER_devel_pf} reports the EER (on the development set) while Tables \ref{AUC_test_pf} and \ref{HTER_test_pf} report AUC (on the test set) and HTER (on the test set). In the tables, "\textit{Spe}" denotes client-specific models whereas "\textit{Ind}" stands for client-independent methods. As can be seen from the tables, the proposed client-specific approach boosts the performance of all one-class anomaly detection methods. The reduction in HTER on the test set can be as high as $\sim 80\%$ for the MD classifier. In terms of AUC on the test set, the margin in the performance improvement can be as high as $\sim 50\%$ for the OCSRC and MD classifiers. The best performing client-specific method in this case is the MD classifier operating on the features extracted via the GoogLeNet model with the HTER of $7.84\%$ and the AUC measure of $97.55\%$ whereas the HTER achieved by the client-independent MD method operating on the same set of deep representations is $38.01\%$ with the AUC of $65.52\%$.

\begin{table}[t]
\centering
\renewcommand{\arraystretch}{1.2}
\caption{EER measures ($\%$) on the development set of the Replay-Attack dataset on a per-frame basis}
\begin{tabular}{c}
\includegraphics[width=2.5in]{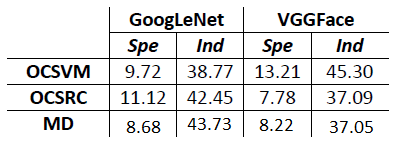}
\end{tabular}
\label{EER_devel_pf}

\centering
\renewcommand{\arraystretch}{1.2}
\caption{AUC measures ($\%$) on the test set of the Replay-Attack dataset on a per-frame basis.}
\begin{tabular}{c}
\includegraphics[width=2.5in]{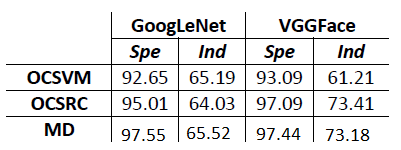}
\end{tabular}
\label{AUC_test_pf}

\centering
\renewcommand{\arraystretch}{1.2}
\caption{HTER measures ($\%$) on the test set of the Replay-Attack dataset on a per-frame basis}
\begin{tabular}{c}
\includegraphics[width=2.5in]{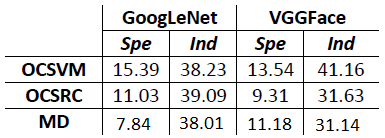}
\end{tabular}
\label{HTER_test_pf}
\end{table}
\subsection{Per-Video Results}
In this section, the performance of the proposed approach is reported on a per-video basis. For this purpose, a score-level fusion approach is applied to obtain the final score for a given video. We opt for the mean fusion rule, averaging the scores of different frames in a given video to produce the final score for the video. As in the per-frame case, the performance of different systems is reported in terms of EER (on the development set), AUC (on the test set) and HTER (on the test set) measures. Tables \ref{EER_devel_pv}, \ref{AUC_test_pv} and \ref{HTER_test_pv} report the EERs, AUCs, and HTERs for the same systems on a per-video basis. 
A notable improvement in performance is observed in the per-video evaluation. As can be verified from the tables, all the client-specific approaches perform better than their client-independent counterparts. On a per-video basis, the largest boost in performance in terms of the AUC measure on the test set corresponds to the OCSVM+VGGFace system with a $\sim 53\%$ improvement. The largest reduction in HTER on the test set again corresponds to the OCSVM+VGGFace system with a $\sim 78\%$ reduction in error rate. The best performing method in this case is the MD+GoogLeNet system with an HTER of $4.62\%$, whereas the best performing client-independent method of MD+VGGFace obtains a HTER of $28.63\%$

\begin{table}[t]
\centering
\renewcommand{\arraystretch}{1.2}
\caption{EER measures ($\%$) on the development set of the Replay-Attack dataset on a per-video basis.}
\begin{tabular}{c}
\includegraphics[width=2.5in]{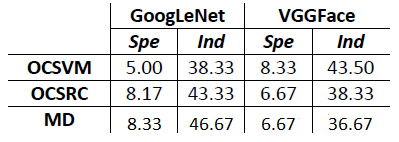}
\end{tabular}
\label{EER_devel_pv}

\centering
\renewcommand{\arraystretch}{1.2}
\caption{AUC measures ($\%$) on the test set of the Replay-Attack dataset on a per-video basis.}
\begin{tabular}{c}
\includegraphics[width=2.5in]{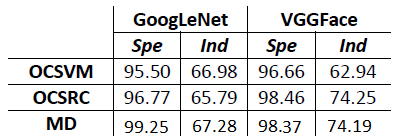}
\end{tabular}
\label{AUC_test_pv}

\centering
\renewcommand{\arraystretch}{1.2}
\caption{HTER measures ($\%$) on the test set of the Replay-Attack dataset on a per-video basis.}
\begin{tabular}{c}
\includegraphics[width=2.5in]{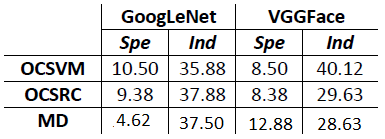}
\end{tabular}
\label{HTER_test_pv}
\end{table}
\begin{figure*}[t]
\centering
\includegraphics[width=3.5in]{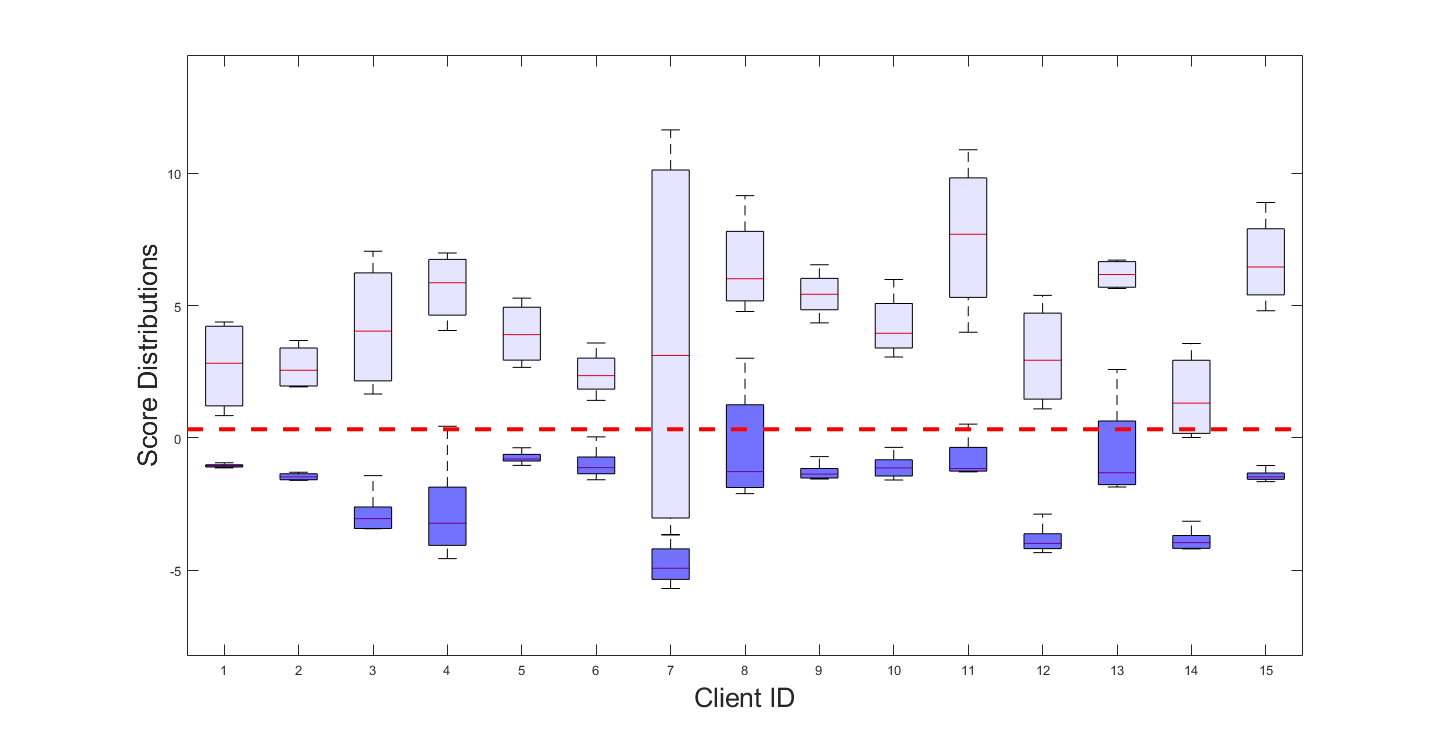}
\includegraphics[width=3.5in]{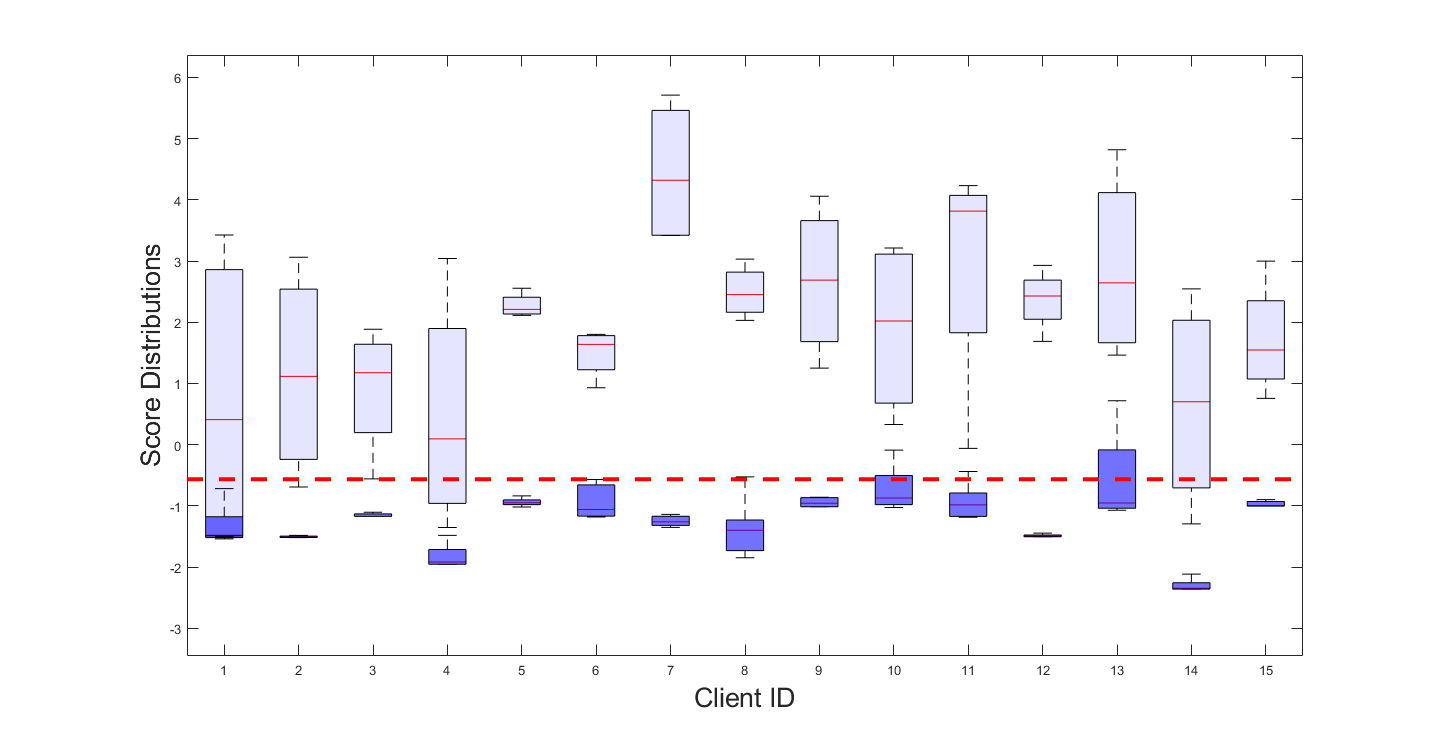}\\
\includegraphics[width=3.5in]{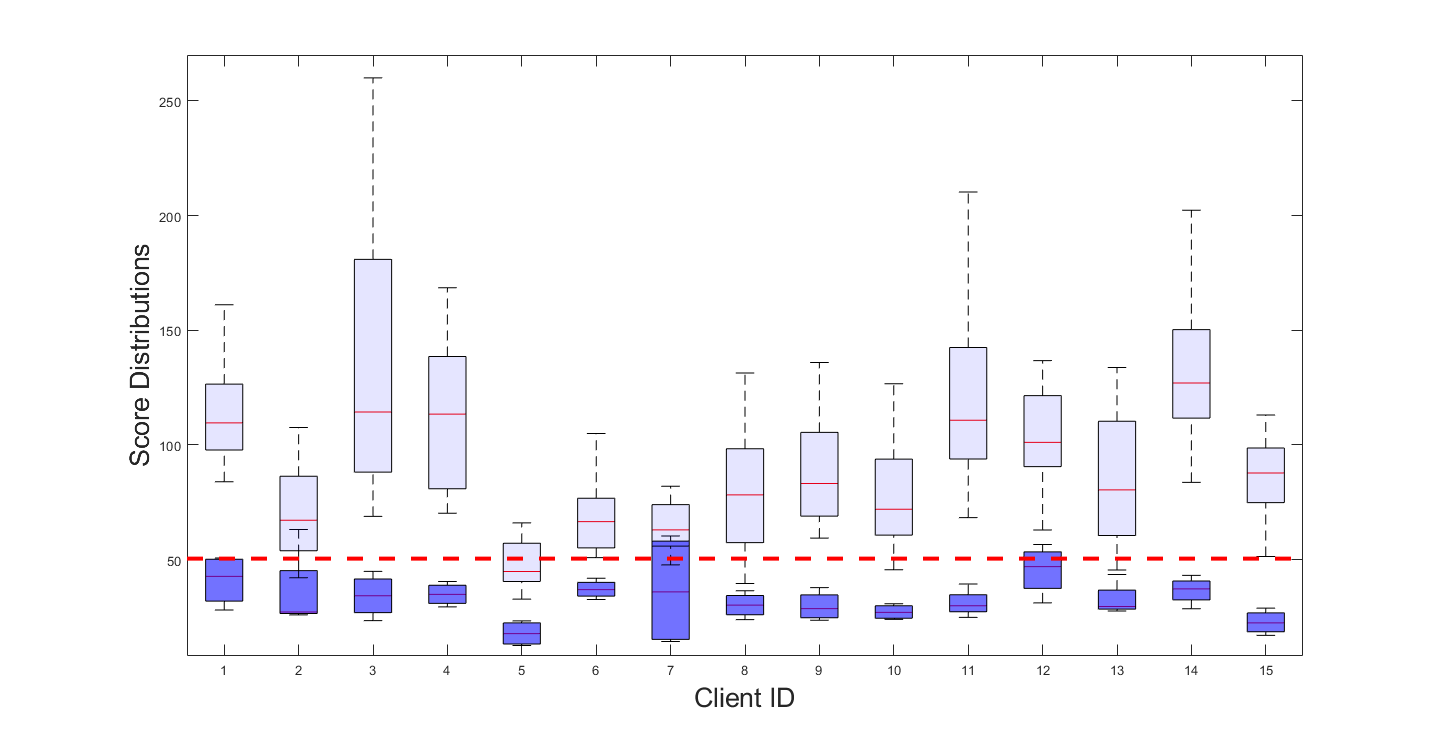}
\includegraphics[width=3.5in]{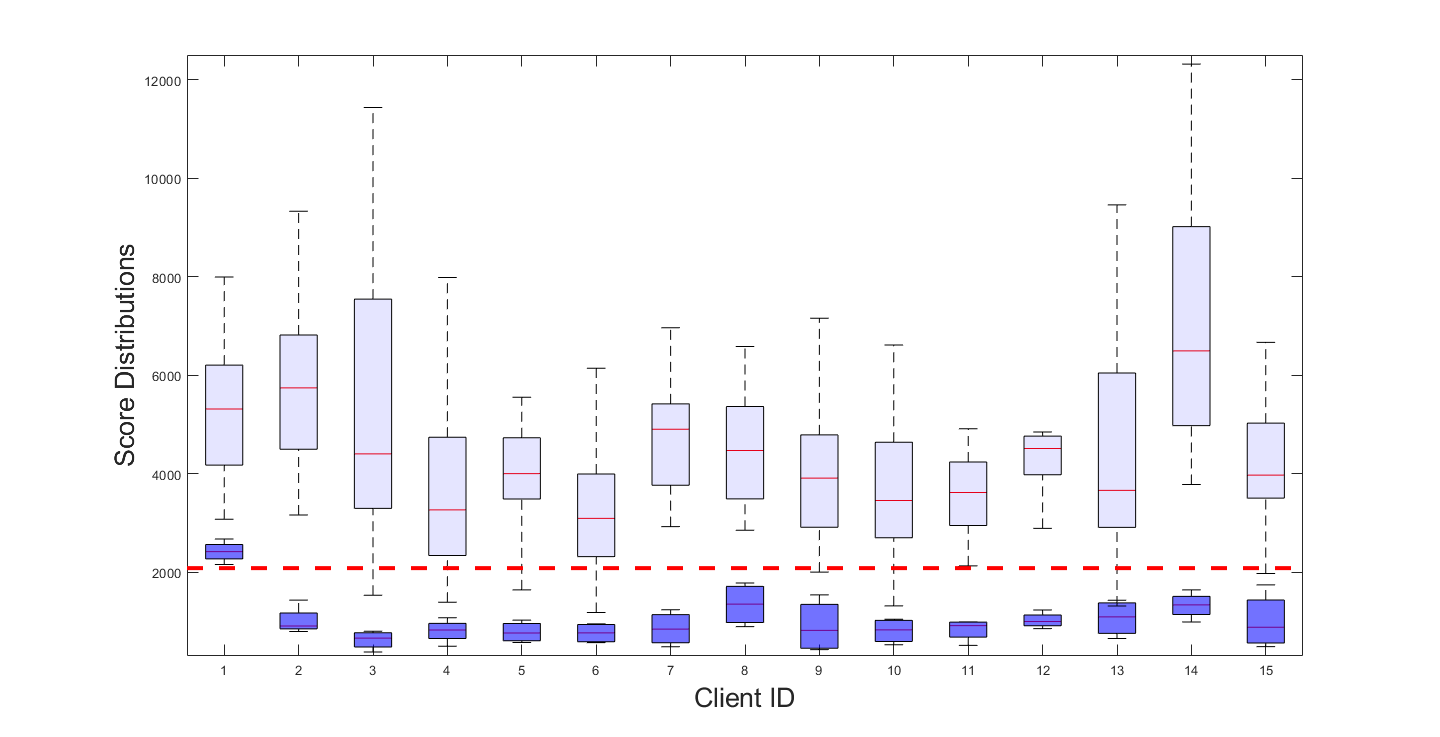}\\
\includegraphics[width=3.5in]{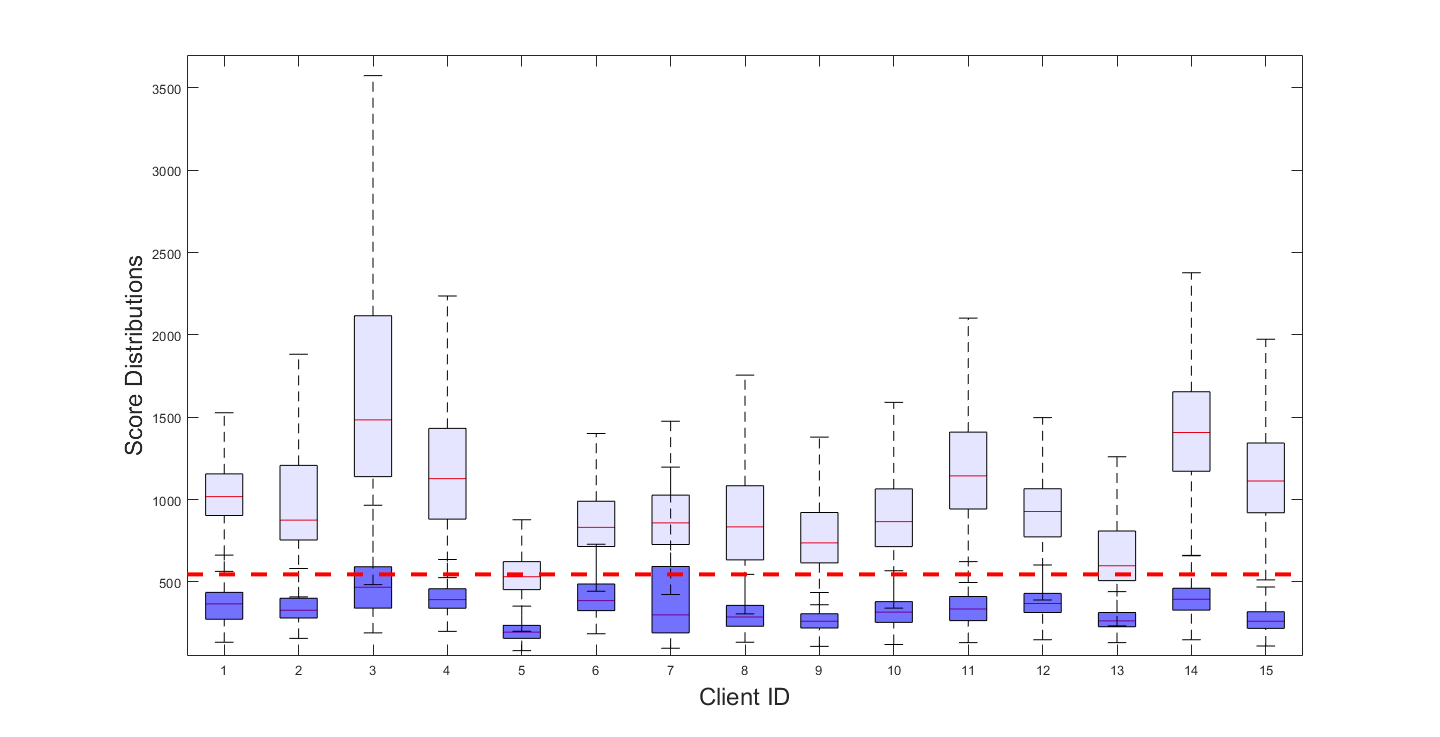}
\includegraphics[width=3.5in]{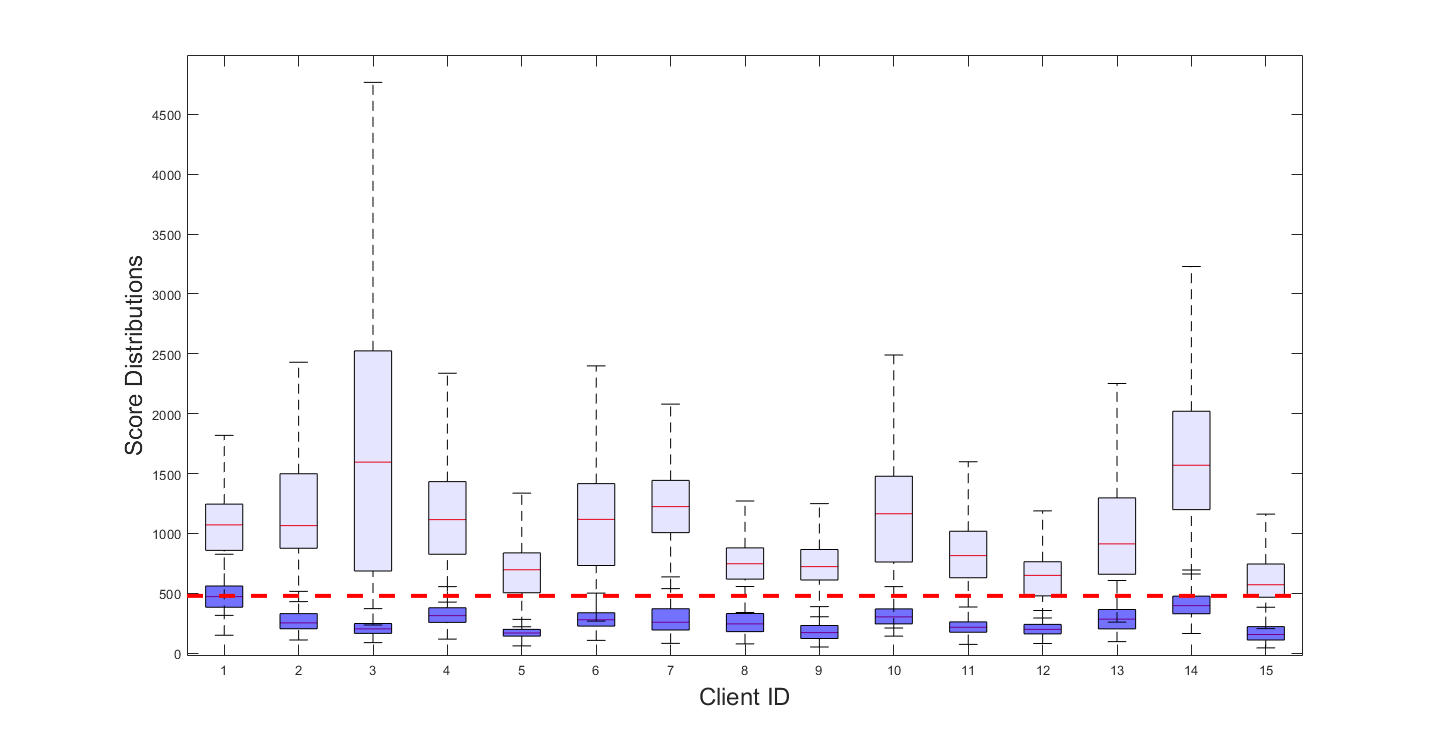}
\caption{Box plots of score distributions of different client-specific systems for the 15 subjects in the development set; From top to bottom in each row: OCSVM, OCSRC and MD classifiers, respectively; Left column GoogLeNet, right column VGGFace}
\label{bps}
\end{figure*}
\subsection{Discussion}
In the experiments conducted so far it is shown that the use of client identity information can result in large improvements in system performance for all the three classifiers examined.

Regarding the deep CNN models examined, the outstanding performance of these networks in the face PAD context illustrates the applicability of such models for face PAD in the one-class client-specific framework. Interestingly, although the GoogLeNet network is not tuned for face recognition, yet it performs reasonably well.
\begin{table*}[t]
\centering
\renewcommand{\arraystretch}{1.2}
\caption{The effect of using a client-specific threshold vs. a global threshold on the EERs ($\%$) of the development set.(G.Thr.: Global Threshold, CS.Thr.: Client-Specific Threshold)}
\begin{tabular}{c}
\includegraphics[width=6in]{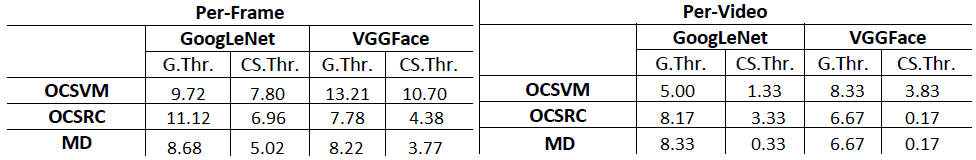}
\end{tabular}
\label{cs-vs-gd}
\end{table*}
\begin{table*}[t]
\centering
\renewcommand{\arraystretch}{1.2}
\caption{The effect of using a client-specific threshold vs. a global threshold on the EERs ($\%$) of the test set.(G.Thr.: Global Threshold, CS.Thr.: Client-Specific Threshold)}
\begin{tabular}{c}
\includegraphics[width=6in]{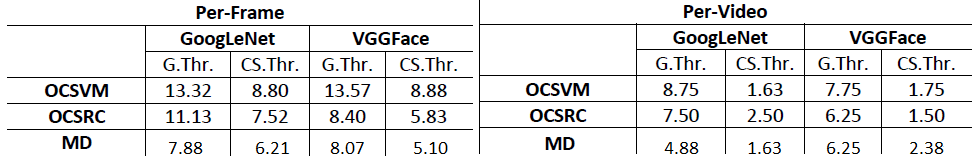}
\end{tabular}
\label{cs-vs-gt}
\end{table*}
\subsection{Class-specific thresholds}
In the preceding experiments, a single global threshold, set on a disjoint set of subjects (the development set), is utilized for all the clients in the decision making step. However, the global threshold could possibly be optimal only if the score distributions corresponding to different subjects were similar. The score distributions for the 15 subjects in the development set for different systems are depicted as box plots in Fig. \ref{bps}, where the horizontal dotted line in each plot corresponds to the global threshold used in the global threshold experiments. As can be seen from the figure, different subjects exhibit different score distributions. As a result, a single global threshold results in a suboptimal performance of the proposed systems.

In order to evaluate the efficacy of using client-specific thresholds, a distinct threshold is set for each subject in each client-specific system and the error rates obtained are compared with those obtained by the systems using a global threshold in Tables \ref{cs-vs-gd} and \ref{cs-vs-gt} on the development and test sets, respectively. As can be seen from the tables, the use of subject-specific thresholds improves the performance of all systems. The best performing method on the test set is the $OCSRC+VGGFace$ with an ERR of $1.50\%$ in the per-video evaluation, where the improvement obtained through the use of subject-specific thresholds reaches $\sim 76\%$. In the per-frame evaluation on the test set, the best performing system is found to be $MD+VGGFace$ with an EER of $5.10\%$, benefiting from the use of client-specific thresholds by a $\sim 37\%$ reduction in EER.
\begin{table*}[t]
\centering
\renewcommand{\arraystretch}{1.2}
\caption{The effect of geometric normalisation based on the eyes coordinates in terms of AUCs ($\%$) on the test set.}
\begin{tabular}{c}
\includegraphics[width=6in]{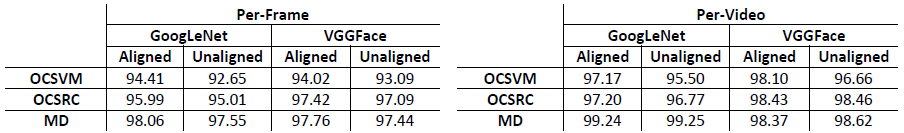}
\end{tabular}
\label{bbox-vs-geo}
\end{table*}
\subsection{Geometric Normalisation}
In the experiments conducted so far, the face bounding box coordinates were used to extract a face from the entire frame. No further geometric pre-processing was applied on the image prior to feature extraction. In this section, the effect of a geometric pre-processing step on the system performance is analysed. To this end, using the method in \cite{6909636} facial landmarks are detected followed by a non-reflective similarity transformation to set the interocular distance to 40 pixels where the size of normalised image is $90\times90$. The rest of the procedure is kept similar to the previous experiments. Table \ref{bbox-vs-geo} reports the corresponding AUCs obtained on the test set on a per-frame and per-video basis and compares these to the case when only bounding boxes are used. As can be observed from the table, a geometric normalisation based on the eyes coordinates moderately improves the system performance for all the evaluated systems in the per-frame evaluation scheme. The largest improvement in AUC's for the per-frame scenario corresponds to the $OCSVM+GoogLeNet_{spe}$ system with a $\sim 1.76\%$ gain. For the per-video evaluation scenario, the largest improvement in performance again corresponds to the $OCSVM+GoogLeN et_{spe}$ system with a $\sim 1.67\%$ gain. It is expected that a more effective and accurate eye detection mechanism other than the method in \cite{6909636} would result in a larger improvement in performance.

\subsection{Training Sample Size}
\begin{figure}[t]
\centering
\includegraphics[width=3.8in]{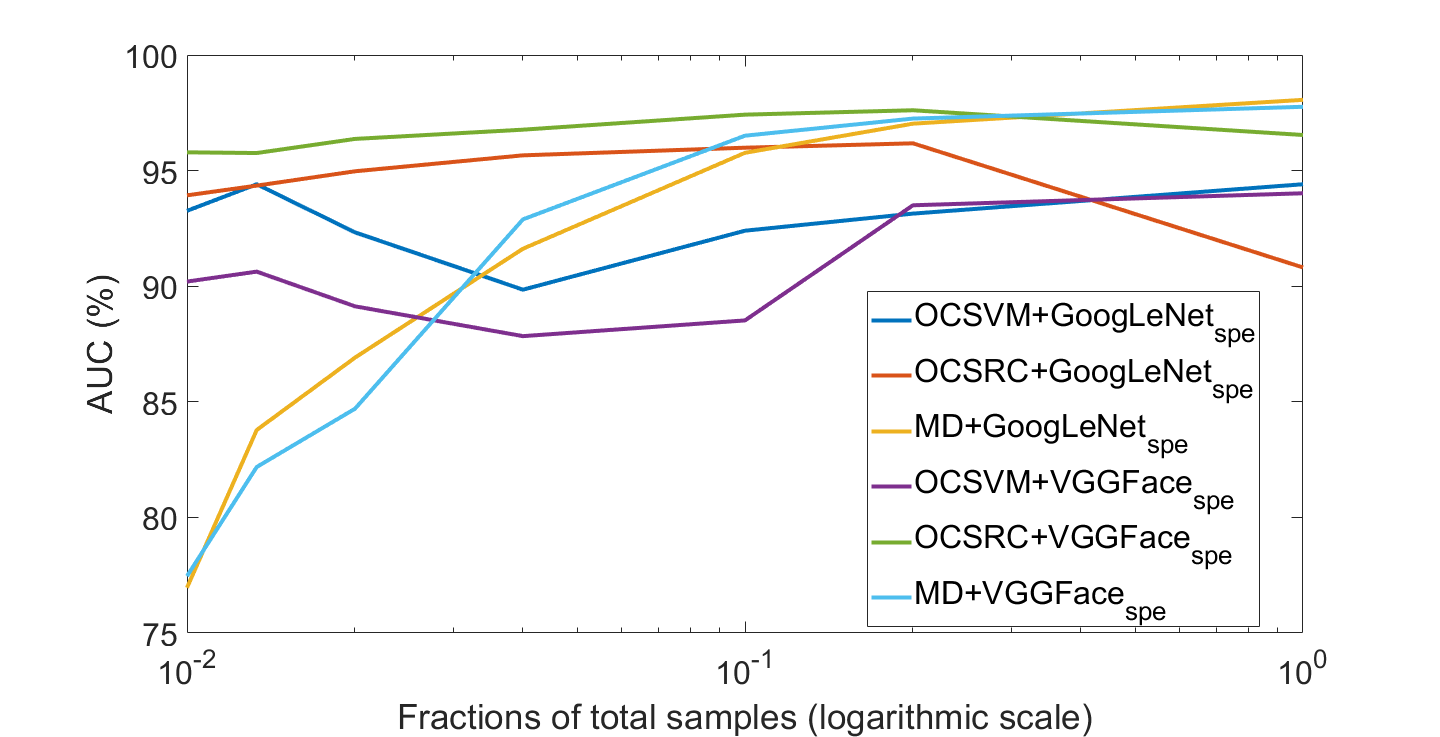}
\caption{The effect of the training sample size on the performance in terms of AUC ($\%$) on the test set.(per-frame evaluation)}
\label{sspf}
\end{figure}

\begin{figure}[t]
\centering
\includegraphics[width=3.8in]{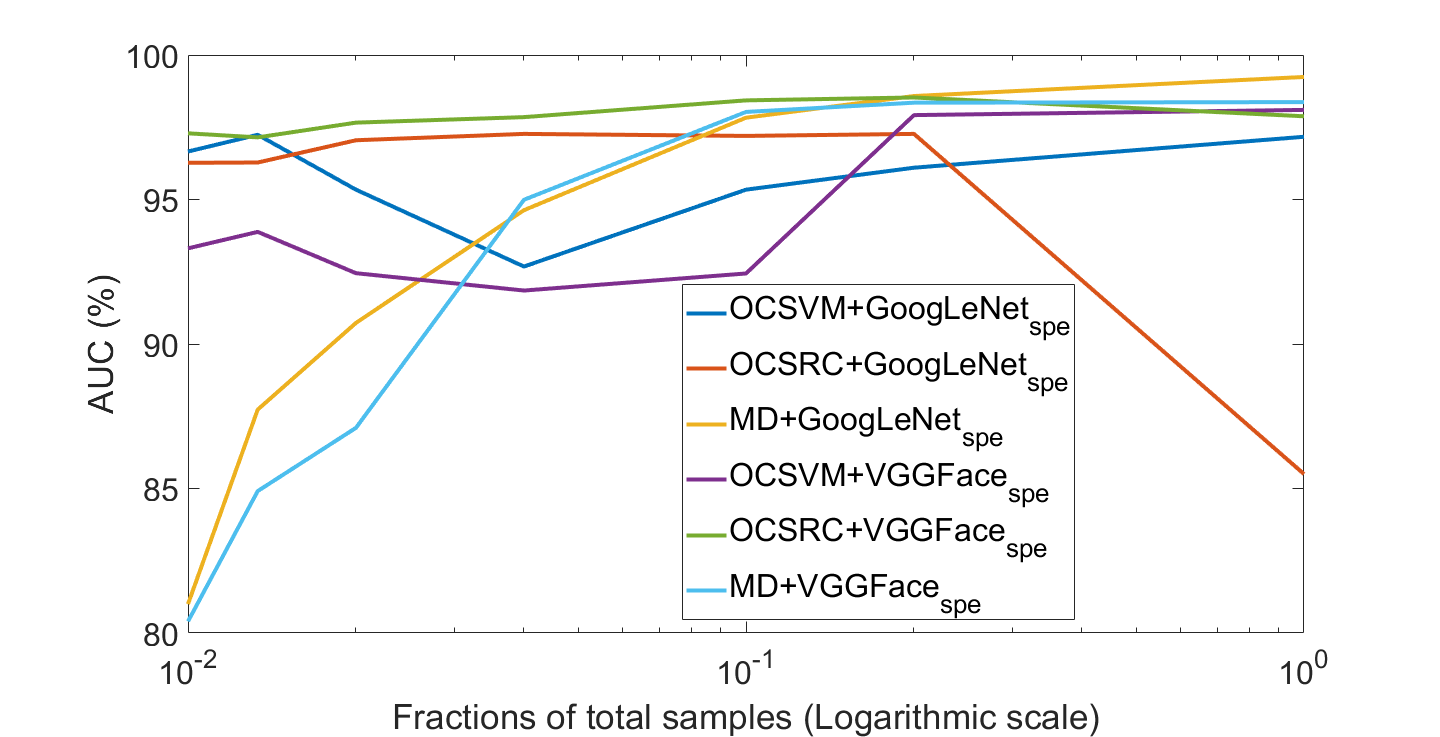}
\caption{The effect of the training sample size on the performance in terms of AUC ($\%$) on the test set.(per-video evaluation)}
\label{sspv}
\end{figure}

In this section, the effect of training sample size on performance is analysed. In this respect, the client-specific one-class methods operating on the aligned version of the test set are considered where the training sample size is gradually increased from $1\%$ of the total samples available to $100\%$. More specifically, the fractions of the total number of available samples used for training the client-specific systems are $\frac{1}{100}$, $\frac{1}{75}$, $\frac{1}{50}$, $\frac{1}{20}$, $\frac{1}{10}$, $\frac{1}{5}$ and $\frac{1}{1}$. Figures \ref{sspf} and \ref{sspv} illustrate the effect of training sample size on the performances of various systems on a per-frame and per-video basis, respectively. As can be observed from the figures, for the systems based on the MD classifier increasing the number of training samples monotonically improves performance. On the other hand, the performance of all systems operating on the one-class SRC basis deteriorates when the training sample size increases from $20\%$ to $100\%$ of the total number of available samples. In this case, while the optimal fraction of the total number of training samples might differ from one system to another, the $10\%$ fraction used in our earlier experiments does not seem to be far from optimal. Regarding the one-class SVM, despite some fluctuations in performance, increasing the number of training samples typically improves performance.   

\subsection{Comparison to Other Methods in the Unseen Scenario}
Despite the fact that there exist different methods to evaluate the generalisation capacity of PAD systems, only a few have followed the \textit{unseen} attack type evaluation protocol. The majority of the methods which have tried a cross-database evaluation are directed more towards evaluating the generalisation capabilities of different systems subject to different imaging conditions (lighting, background, sensor-interoperability, etc.) rather than attack type. Among other two-class systems, the work in \cite{7820995} and \cite{7031941} have followed the \textit{unseen} evaluation scheme on the Replay-Attack dataset. While the evaluation scheme considered in \cite{7820995} is unseen in the sense that it excludes one of the three attack types (Print, Digital Photo and Video) during training in each of the considered scenarios, the evaluation cannot be considered completely unseen as the authors use similar attack types (video replays) both for training and evaluation in some of the their evaluations. As a result, our comparison is limited to the work in \cite{7031941,Chingovska_THESIS_2015}. As the subjects in the development and test set of the Replay-Attack dataset are disjoint and setting a client-specific threshold requires development data for each test subject, it is impossible to assess the performance of the proposed client-specific approach using subject-specific thresholds in terms of the usual HTER measure on the test set. As a result, the comparison is limited to the EER measures on the development set. The best EER on the development set obtained using a discriminative approach in \cite{Chingovska_THESIS_2015} is $2.83\%$, whereas the best EER on the development set obtained in this work is $0.17\%$ (face bounding boxes) obtained for both the $OCSRC+VGGFace_{spe}$ and $MD+VGGFace_{spe}$ systems with client-specific thresholds, in spite of the fact that no attack data has been used for training.

\section{Conclusion}
\label{conc}
A novel approach to face presentation attack detection in the unseen attack scenario is developed. Motivated by the promising performance of the one-class anomaly detection approaches, a client-specific version of the one-class methodology is proposed for the detection of face presentation attacks.  Both generative and discriminative one-class classifiers utilising only positive samples (real access data) for training are examined. It is shown that the use of client identity information in the model construction can boost the system performance of both discriminative and generative approaches significantly. Based on the score distributions of different clients, subject-specific thresholds are determined and used further to improve the performance. Different deep CNN models serving as mechanisms for feature extraction in face PAD have been evaluated. It has been shown that the same set of deep CNN features as that used for face recognition can be employed  for the presentation attack detection in the context of the proposed client-specific one-class approach. A comparison of the proposed one-class client-specific approaches to  two-class methods in the unseen attack scenario confirmed the merits of the proposed approach.

\section*{Acknowledgement}

\ifCLASSOPTIONcaptionsoff
  \newpage
\fi



%
\bibliographystyle{IEEEtran}
\bibliography{IEEEexample}

\begin{thebibliography}{10}
\providecommand{\url}[1]{#1}
\csname url@samestyle\endcsname
\providecommand{\newblock}{\relax}
\providecommand{\bibinfo}[2]{#2}
\providecommand{\BIBentrySTDinterwordspacing}{\spaceskip=0pt\relax}
\providecommand{\BIBentryALTinterwordstretchfactor}{4}
\providecommand{\BIBentryALTinterwordspacing}{\spaceskip=\fontdimen2\font plus
\BIBentryALTinterwordstretchfactor\fontdimen3\font minus
  \fontdimen4\font\relax}
\providecommand{\BIBforeignlanguage}[2]{{%
\expandafter\ifx\csname l@#1\endcsname\relax
\typeout{** WARNING: IEEEtran.bst: No hyphenation pattern has been}%
\typeout{** loaded for the language `#1'. Using the pattern for}%
\typeout{** the default language instead.}%
\else
\language=\csname l@#1\endcsname
\fi
#2}}
\providecommand{\BIBdecl}{\relax}
\BIBdecl

\bibitem{7984788}
S.~R. Arashloo, J.~Kittler, and W.~Christmas, ``An anomaly detection approach
  to face spoofing detection: A new formulation and evaluation protocol,''
  \emph{IEEE Access}, vol.~5, pp. 13\,868--13\,882, 2017.

\bibitem{Nikisins_ICB2018_2018/LIDIAP}
O.~Nikisins, A.~Mohammadi, A.~Anjos, and S.~Marcel, ``On {E}ffectiveness of
  {A}nomaly {D}etection {A}pproaches against {U}nseen {P}resentation {A}ttacks
  in {F}ace {A}nti-{S}poofing,'' in \emph{The 11th {IAPR} {I}nternational
  {C}onference on {B}iometrics ({ICB} 2018)}, 2018.

\bibitem{7031941}
I.~Chingovska and A.~R. dos Anjos, ``On the use of client identity information
  for face antispoofing,'' \emph{IEEE Transactions on Information Forensics and
  Security}, vol.~10, no.~4, pp. 787--796, April 2015.

\bibitem{7041231}
J.~Yang, Z.~Lei, D.~Yi, and S.~Z. Li, ``Person-specific face antispoofing with
  subject domain adaptation,'' \emph{IEEE Transactions on Information Forensics
  and Security}, vol.~10, no.~4, pp. 797--809, April 2015.

\bibitem{7820995}
T.~Edmunds and A.~Caplier, ``Fake face detection based on radiometric
  distortions,'' in \emph{2016 Sixth International Conference on Image
  Processing Theory, Tools and Applications (IPTA)}, Dec 2016, pp. 1--6.

\bibitem{6905848}
S.~R. Arashloo and J.~Kittler, ``Class-specific kernel fusion of multiple
  descriptors for face verification using multiscale binarised statistical
  image features,'' \emph{IEEE Transactions on Information Forensics and
  Security}, vol.~9, no.~12, pp. 2100--2109, Dec 2014.

\bibitem{Arashloo2017}
\BIBentryALTinterwordspacing
S.~R. Arashloo, ``Multiscale binarised statistical image features for symmetric
  face matching using multiple descriptor fusion based on class-specific lda,''
  \emph{Pattern Analysis and Applications}, vol.~20, no.~1, pp. 113--126, Feb
  2017. [Online]. Available: \url{https://doi.org/10.1007/s10044-015-0475-1}
\BIBentrySTDinterwordspacing

\bibitem{7494969}
A.~Iosifidis and M.~Gabbouj, ``Scaling up class-specific kernel discriminant
  analysis for large-scale face verification,'' \emph{IEEE Transactions on
  Information Forensics and Security}, vol.~11, no.~11, pp. 2453--2465, Nov
  2016.

\bibitem{THANHTRAN2017131}
D.~T. Tran, M.~Gabbouj, and A.~Iosifidis, ``Multilinear class-specific
  discriminant analysis,'' \emph{Pattern Recognition Letters}, vol. 100, no.
  Supplement C, pp. 131 -- 136, 2017.

\bibitem{7965980}
A.~Iosifidis and M.~Gabbouj, ``Class-specific kernel discriminant analysis
  based on cholesky decomposition,'' in \emph{2017 International Joint
  Conference on Neural Networks (IJCNN)}, May 2017, pp. 1141--1146.

\bibitem{iet:/content/journals/10.1049/iet-bmt.2017.0081}
G.~Cao, ``\BIBforeignlanguage{English}{Neural class-specific regression for
  face verification},'' \emph{\BIBforeignlanguage{English}{IET Biometrics}},
  vol.~7, pp. 63--70(7), January 2018.

\bibitem{edmunds:tel-01576830}
\BIBentryALTinterwordspacing
T.~Edmunds, ``{Protection of 2D face identification systems against spoofing
  attacks.}'' Theses, {Univ. Grenoble Alpes}, Jan. 2017. [Online]. Available:
  \url{https://hal.archives-ouvertes.fr/tel-01576830}
\BIBentrySTDinterwordspacing

\bibitem{6612968}
J.~Komulainen, A.~Hadid, M.~Pietikäinen, A.~Anjos, and S.~Marcel,
  ``Complementary countermeasures for detecting scenic face spoofing attacks,''
  in \emph{2013 International Conference on Biometrics (ICB)}, June 2013, pp.
  1--7.

\bibitem{6313548}
I.~Chingovska, A.~Anjos, and S.~Marcel, ``On the effectiveness of local binary
  patterns in face anti-spoofing,'' in \emph{2012 BIOSIG - Proceedings of the
  International Conference of Biometrics Special Interest Group (BIOSIG)}, Sept
  2012, pp. 1--7.

\bibitem{6117592}
W.~R. Schwartz, A.~Rocha, and H.~Pedrini, ``Face spoofing detection through
  partial least squares and low-level descriptors,'' in \emph{2011
  International Joint Conference on Biometrics (IJCB)}, Oct 2011, pp. 1--8.

\bibitem{6847399}
K.~B. Housam, S.~H. Lau, Y.~H. Pang, Y.~P. Liew, and M.~L. Chiang, ``Face
  spoofing detection based on improved local graph structure,'' in \emph{2014
  International Conference on Information Science Applications (ICISA)}, May
  2014, pp. 1--4.

\bibitem{7163625}
S.~R. Arashloo, J.~Kittler, and W.~Christmas, ``Face spoofing detection based
  on multiple descriptor fusion using multiscale dynamic binarized statistical
  image features,'' \emph{IEEE Transactions on Information Forensics and
  Security}, vol.~10, no.~11, pp. 2396--2407, Nov 2015.

\bibitem{6612955}
J.~Yang, Z.~Lei, S.~Liao, and S.~Z. Li, ``Face liveness detection with
  component dependent descriptor,'' in \emph{2013 International Conference on
  Biometrics (ICB)}, June 2013, pp. 1--6.

\bibitem{4409068}
G.~Pan, L.~Sun, Z.~Wu, and S.~Lao, ``Eyeblink-based anti-spoofing in face
  recognition from a generic webcamera,'' in \emph{2007 IEEE 11th International
  Conference on Computer Vision}, Oct 2007, pp. 1--8.

\bibitem{4563115}
K.~Kollreider, H.~Fronthaler, and J.~Bigun, ``Verifying liveness by multiple
  experts in face biometrics,'' in \emph{2008 IEEE Computer Society Conference
  on Computer Vision and Pattern Recognition Workshops}, June 2008, pp. 1--6.

\bibitem{KOLLREIDER2009233}
\BIBentryALTinterwordspacing
------, ``Non-intrusive liveness detection by face images,'' \emph{Image and
  Vision Computing}, vol.~27, no.~3, pp. 233 -- 244, 2009, special Issue on
  Multimodal Biometrics. [Online]. Available:
  \url{http://www.sciencedirect.com/science/article/pii/S0262885607000893}
\BIBentrySTDinterwordspacing

\bibitem{FENG2016451}
\BIBentryALTinterwordspacing
L.~Feng, L.-M. Po, Y.~Li, X.~Xu, F.~Yuan, T.~C.-H. Cheung, and K.-W. Cheung,
  ``Integration of image quality and motion cues for face anti-spoofing: A
  neural network approach,'' \emph{Journal of Visual Communication and Image
  Representation}, vol.~38, pp. 451 -- 460, 2016. [Online]. Available:
  \url{http://www.sciencedirect.com/science/article/pii/S1047320316300244}
\BIBentrySTDinterwordspacing

\bibitem{6485156}
J.~Yan, Z.~Zhang, Z.~Lei, D.~Yi, and S.~Z. Li, ``Face liveness detection by
  exploring multiple scenic clues,'' in \emph{2012 12th International
  Conference on Control Automation Robotics Vision (ICARCV)}, Dec 2012, pp.
  188--193.

\bibitem{7185398}
A.~Pinto, H.~Pedrini, W.~R. Schwartz, and A.~Rocha, ``Face spoofing detection
  through visual codebooks of spectral temporal cubes,'' \emph{IEEE
  Transactions on Image Processing}, vol.~24, no.~12, pp. 4726--4740, Dec 2015.

\bibitem{6199760}
G.~Kim, S.~Eum, J.~K. Suhr, D.~I. Kim, K.~R. Park, and J.~Kim, ``Face liveness
  detection based on texture and frequency analyses,'' in \emph{2012 5th IAPR
  International Conference on Biometrics (ICB)}, March 2012, pp. 67--72.

\bibitem{6382760}
A.~d.~S.~Pinto, H.~Pedrini, W.~Schwartz, and A.~Rocha, ``Video-based face
  spoofing detection through visual rhythm analysis,'' in \emph{2012 25th
  SIBGRAPI Conference on Graphics, Patterns and Images}, Aug 2012, pp.
  221--228.

\bibitem{7017526}
A.~Pinto, W.~R. Schwartz, H.~Pedrini, and A.~d.~R.~Rocha, ``Using visual
  rhythms for detecting video-based facial spoof attacks,'' \emph{IEEE
  Transactions on Information Forensics and Security}, vol.~10, no.~5, pp.
  1025--1038, May 2015.

\bibitem{6976921}
J.~Galbally and S.~Marcel, ``Face anti-spoofing based on general image quality
  assessment,'' in \emph{2014 22nd International Conference on Pattern
  Recognition}, Aug 2014, pp. 1173--1178.

\bibitem{7351280}
Z.~Boulkenafet, J.~Komulainen, and A.~Hadid, ``Face anti-spoofing based on
  color texture analysis,'' in \emph{2015 IEEE International Conference on
  Image Processing (ICIP)}, Sept 2015, pp. 2636--2640.

\bibitem{6612957}
T.~Wang, J.~Yang, Z.~Lei, S.~Liao, and S.~Z. Li, ``Face liveness detection
  using 3d structure recovered from a single camera,'' in \emph{2013
  International Conference on Biometrics (ICB)}, June 2013, pp. 1--6.

\bibitem{6622704}
N.~Kose and J.~L. Dugelay, ``Reflectance analysis based countermeasure
  technique to detect face mask attacks,'' in \emph{2013 18th International
  Conference on Digital Signal Processing (DSP)}, July 2013, pp. 1--6.

\bibitem{KOSE2014779}
\BIBentryALTinterwordspacing
N.~Kose and J.-L. Dugelay, ``Mask spoofing in face recognition and
  countermeasures,'' \emph{Image and Vision Computing}, vol.~32, no.~10, pp.
  779 -- 789, 2014, best of Automatic Face and Gesture Recognition 2013.
  [Online]. Available:
  \url{http://www.sciencedirect.com/science/article/pii/S0262885614001024}
\BIBentrySTDinterwordspacing

\bibitem{10.1007/978-3-642-15567-3_37}
X.~Tan, Y.~Li, J.~Liu, and L.~Jiang, ``Face liveness detection from a single
  image with sparse low rank bilinear discriminative model,'' in \emph{Computer
  Vision -- ECCV 2010}, K.~Daniilidis, P.~Maragos, and N.~Paragios, Eds.\hskip
  1em plus 0.5em minus 0.4em\relax Berlin, Heidelberg: Springer Berlin
  Heidelberg, 2010, pp. 504--517.

\bibitem{6117510}
J.~Määttä, A.~Hadid, and M.~Pietikäinen, ``Face spoofing detection from
  single images using micro-texture analysis,'' in \emph{2011 International
  Joint Conference on Biometrics (IJCB)}, Oct 2011, pp. 1--7.

\bibitem{6712690}
J.~Komulainen, A.~Hadid, and M.~Pietikäinen, ``Context based face
  anti-spoofing,'' in \emph{2013 IEEE Sixth International Conference on
  Biometrics: Theory, Applications and Systems (BTAS)}, Sept 2013, pp. 1--8.

\bibitem{6180283}
J.~Maatta, A.~Hadid, and M.~Pietikainen, ``Face spoofing detection from single
  images using texture and local shape analysis,'' \emph{IET Biometrics},
  vol.~1, no.~1, pp. 3--10, March 2012.

\bibitem{10.1007/978-3-642-37410-4_11}
T.~de~Freitas~Pereira, A.~Anjos, J.~M. De~Martino, and S.~Marcel,
  ``Lbp{\thinspace}−{\thinspace}top based countermeasure against face
  spoofing attacks,'' in \emph{Computer Vision - ACCV 2012 Workshops}, J.-I.
  Park and J.~Kim, Eds.\hskip 1em plus 0.5em minus 0.4em\relax Berlin,
  Heidelberg: Springer Berlin Heidelberg, 2013, pp. 121--132.

\bibitem{7031384}
D.~Wen, H.~Han, and A.~K. Jain, ``Face spoof detection with image distortion
  analysis,'' \emph{IEEE Transactions on Information Forensics and Security},
  vol.~10, no.~4, pp. 746--761, April 2015.

\bibitem{6595861}
S.~Bharadwaj, T.~I. Dhamecha, M.~Vatsa, and R.~Singh, ``Computationally
  efficient face spoofing detection with motion magnification,'' in \emph{2013
  IEEE Conference on Computer Vision and Pattern Recognition Workshops}, June
  2013, pp. 105--110.

\bibitem{6810829}
N.~Erdogmus and S.~Marcel, ``Spoofing face recognition with 3d masks,''
  \emph{IEEE Transactions on Information Forensics and Security}, vol.~9,
  no.~7, pp. 1084--1097, July 2014.

\bibitem{7029061}
D.~Menotti, G.~Chiachia, A.~Pinto, W.~R. Schwartz, H.~Pedrini, A.~X. Falcão,
  and A.~Rocha, ``Deep representations for iris, face, and fingerprint spoofing
  detection,'' \emph{IEEE Transactions on Information Forensics and Security},
  vol.~10, no.~4, pp. 864--879, April 2015.

\bibitem{6117509}
M.~M. Chakka, A.~Anjos, S.~Marcel, R.~Tronci, D.~Muntoni, G.~Fadda, M.~Pili,
  N.~Sirena, G.~Murgia, M.~Ristori, F.~Roli, J.~Yan, D.~Yi, Z.~Lei, Z.~Zhang,
  S.~Z. Li, W.~R. Schwartz, A.~Rocha, H.~Pedrini, J.~Lorenzo-Navarro,
  M.~Castrillón-Santana, J.~Määttä, A.~Hadid, and M.~Pietikäinen,
  ``Competition on counter measures to 2-d facial spoofing attacks,'' in
  \emph{2011 International Joint Conference on Biometrics (IJCB)}, Oct 2011,
  pp. 1--6.

\bibitem{6613026}
I.~Chingovska, J.~Yang, Z.~Lei, D.~Yi, S.~Z. Li, O.~Kahm, C.~Glaser, N.~Damer,
  A.~Kuijper, A.~Nouak, J.~Komulainen, T.~Pereira, S.~Gupta, S.~K. Wa,
  S.~Bansal, A.~Rai, T.~Krishna, D.~Goyal, M.~A. Waris, H.~Zhang, I.~Ahmad,
  S.~Kiranyaz, M.~Gabbouj, R.~Tronci, M.~Pili, N.~Sirena, F.~Roli, J.~Galbally,
  J.~Ficrrcz, A.~Pinto, H.~Pedrini, W.~S. Schwartz, A.~Rocha, A.~Anjos, and
  S.~Marcel, ``The 2nd competition on counter measures to 2d face spoofing
  attacks,'' in \emph{2013 International Conference on Biometrics (ICB)}, June
  2013, pp. 1--6.

\bibitem{6116484}
B.~Peixoto, C.~Michelassi, and A.~Rocha, ``Face liveness detection under bad
  illumination conditions,'' in \emph{2011 18th IEEE International Conference
  on Image Processing}, Sept 2011, pp. 3557--3560.

\bibitem{10.1007/978-3-642-21605-3_19}
E.~E.~A. Abusham and H.~K. Bashir, ``Face recognition using local graph
  structure (lgs),'' in \emph{Human-Computer Interaction. Interaction
  Techniques and Environments}, J.~A. Jacko, Ed.\hskip 1em plus 0.5em minus
  0.4em\relax Berlin, Heidelberg: Springer Berlin Heidelberg, 2011, pp.
  169--175.

\bibitem{6317336}
N.~Kose and J.~L. Dugelay, ``Classification of captured and recaptured images
  to detect photograph spoofing,'' in \emph{2012 International Conference on
  Informatics, Electronics Vision (ICIEV)}, May 2012, pp. 1027--1032.

\bibitem{7056504}
D.~C. Garcia and R.~L. de~Queiroz, ``Face-spoofing 2d-detection based on
  moire-pattern analysis,'' \emph{IEEE Transactions on Information Forensics
  and Security}, vol.~10, no.~4, pp. 778--786, April 2015.

\bibitem{6909967}
A.~Hadid, ``Face biometrics under spoofing attacks: Vulnerabilities,
  countermeasures, open issues, and research directions,'' in \emph{2014 IEEE
  Conference on Computer Vision and Pattern Recognition Workshops}, June 2014,
  pp. 113--118.

\bibitem{6636290}
J.~Kittler, W.~Christmas, T.~de~Campos, D.~Windridge, F.~Yan, J.~Illingworth,
  and M.~Osman, ``Domain anomaly detection in machine perception: A system
  architecture and taxonomy,'' \emph{IEEE Transactions on Pattern Analysis and
  Machine Intelligence}, vol.~36, no.~5, pp. 845--859, May 2014.

\bibitem{1262509}
F.~Esponda, S.~Forrest, and P.~Helman, ``A formal framework for positive and
  negative detection schemes,'' \emph{IEEE Transactions on Systems, Man, and
  Cybernetics, Part B (Cybernetics)}, vol.~34, no.~1, pp. 357--373, Feb 2004.

\bibitem{doi:10.1162/089976601750264965}
\BIBentryALTinterwordspacing
B.~Schölkopf, J.~C. Platt, J.~Shawe-Taylor, A.~J. Smola, and R.~C. Williamson,
  ``Estimating the support of a high-dimensional distribution,'' \emph{Neural
  Computation}, vol.~13, no.~7, pp. 1443--1471, 2001. [Online]. Available:
  \url{https://doi.org/10.1162/089976601750264965}
\BIBentrySTDinterwordspacing

\bibitem{4483511}
J.~Wright, A.~Y. Yang, A.~Ganesh, S.~S. Sastry, and Y.~Ma, ``Robust face
  recognition via sparse representation,'' \emph{IEEE Transactions on Pattern
  Analysis and Machine Intelligence}, vol.~31, no.~2, pp. 210--227, Feb 2009.

\bibitem{7102696}
Z.~Zhang, Y.~Xu, J.~Yang, X.~Li, and D.~Zhang, ``A survey of sparse
  representation: Algorithms and applications,'' \emph{IEEE Access}, vol.~3,
  pp. 490--530, 2015.

\bibitem{4655448}
D.~L. Donoho and Y.~Tsaig, ``Fast solution of $\ell _{1}$ -norm minimization
  problems when the solution may be sparse,'' \emph{IEEE Transactions on
  Information Theory}, vol.~54, no.~11, pp. 4789--4812, Nov 2008.

\bibitem{doi:10.1093/imanum/20.3.389}
\BIBentryALTinterwordspacing
M.~Osborne, B.~Presnell, and B.~Turlach, ``A new approach to variable selection
  in least squares problems,'' \emph{IMA Journal of Numerical Analysis},
  vol.~20, no.~3, pp. 389--403, 2000. [Online]. Available: \url{+
  http://dx.doi.org/10.1093/imanum/20.3.389}
\BIBentrySTDinterwordspacing

\bibitem{7298594}
C.~Szegedy, W.~Liu, Y.~Jia, P.~Sermanet, S.~Reed, D.~Anguelov, D.~Erhan,
  V.~Vanhoucke, and A.~Rabinovich, ``Going deeper with convolutions,'' in
  \emph{2015 IEEE Conference on Computer Vision and Pattern Recognition
  (CVPR)}, June 2015, pp. 1--9.

\bibitem{ILSVRC15}
O.~Russakovsky, J.~Deng, H.~Su, J.~Krause, S.~Satheesh, S.~Ma, Z.~Huang,
  A.~Karpathy, A.~Khosla, M.~Bernstein, A.~C. Berg, and L.~Fei-Fei, ``{ImageNet
  Large Scale Visual Recognition Challenge},'' \emph{International Journal of
  Computer Vision (IJCV)}, vol. 115, no.~3, pp. 211--252, 2015.

\bibitem{Parkhi15}
O.~M. Parkhi, A.~Vedaldi, and A.~Zisserman, ``Deep face recognition,'' in
  \emph{British Machine Vision Conference}, 2015.

\bibitem{LFWTech}
G.~B. Huang, M.~Ramesh, T.~Berg, and E.~Learned-Miller, ``Labeled faces in the
  wild: A database for studying face recognition in unconstrained
  environments,'' University of Massachusetts, Amherst, Tech. Rep. 07-49,
  October 2007.

\bibitem{5995566}
L.~Wolf, T.~Hassner, and I.~Maoz, ``Face recognition in unconstrained videos
  with matched background similarity,'' in \emph{CVPR 2011}, June 2011, pp.
  529--534.

\bibitem{6199754}
Z.~Zhang, J.~Yan, S.~Liu, Z.~Lei, D.~Yi, and S.~Z. Li, ``A face antispoofing
  database with diverse attacks,'' in \emph{2012 5th IAPR International
  Conference on Biometrics (ICB)}, March 2012, pp. 26--31.

\bibitem{6909636}
A.~Asthana, S.~Zafeiriou, S.~Cheng, and M.~Pantic, ``Incremental face alignment
  in the wild,'' in \emph{2014 IEEE Conference on Computer Vision and Pattern
  Recognition}, June 2014, pp. 1859--1866.

\bibitem{Chingovska_THESIS_2015}
\BIBentryALTinterwordspacing
I.~Chingovska, ``Trustworthy biometric verification under spoofing attacks:
  Application to the face mode,'' Ph.D. dissertation, {\'{E}}cole Polytechnique
  F{\'{e}}d{\'{e}}rale de Lausanne, Nov. 2015, th{\`{e}}se EPFL, n° 6879
  (2016). [Online]. Available:
  \url{https://pypi.python.org/pypi/bob.thesis.ichingo2015/0.0.1}
\BIBentrySTDinterwordspacing

\end{thebibliography}

%





\end{document}